\documentclass[11pt,a4paper]{article}
\usepackage[hyperref]{naaclhlt2018}
\usepackage{latexsym}

\usepackage{url}

\aclfinalcopy 
\def\aclpaperid{445} 

\usepackage{mathptmx}

\usepackage{comment}

\usepackage{lipsum}
\usepackage{color}
\usepackage{graphicx}

\def\parsum#1{\bgroup \textcolor{blue}{Paragraph summary: #1}\egroup}
\def\sectionsum#1{\bgroup \textcolor{green}{Section content: #1}\egroup \\}

\newif\ifarxiv
\arxivtrue

\title{Before Name-calling: Dynamics and Triggers of Ad Hominem Fallacies in~Web Argumentation}

\author{
Ivan Habernal$^{\dagger}$ \quad Henning Wachsmuth$^{\ddagger}$ \quad Iryna Gurevych$^{\dagger}$ \quad Benno Stein$^{\ddagger}$ \\[.3em]
$^\dagger$ Ubiquitous Knowledge Processing Lab (UKP) and Research Training Group AIPHES  \\
Department of Computer Science, Technische Universit\"{a}t Darmstadt, Germany\\
{\tt www.ukp.tu-darmstadt.de} \quad {\tt www.aiphes.tu-darmstadt.de} \\
$^{\ddagger}$ Faculty of Media, Bauhaus-Universit\"{a}t Weimar, Germany \\
{\tt <firstname>.<lastname>@uni-weimar.de}
}

\date{}

\begin{document}

\ifarxiv
\onecolumn
\noindent \textbf{Before Name-calling: Dynamics and Triggers of Ad Hominem Fallacies in~Web Argumentation}

\medskip
\noindent Ivan Habernal, Henning Wachsmuth, Iryna Gurevych, Benno Stein

\bigskip
This is a \textbf{pre-print non-final version} of the article accepted for publication at the \emph{2018 Conference of the North American Chapter of the Association for Computational Linguistics: Human Language Technologies (NAACL 2018)}. The final official version along with the supplementary materials will be published on the ACL Anthology website in June 2018: \url{http://aclweb.org/anthology/}

\medskip
Please cite this pre-print version as follows.
\medskip

\begin{verbatim}
@InProceedings{habernal.et.al.2018.NAACL.adhominem,
  title = {Before Name-calling: Dynamics and Triggers of Ad Hominem
           Fallacies in Web Argumentation},
  author = {Habernal, Ivan and Wachsmuth, Henning and
            Gurevych, Iryna and Stein, Benno},
  publisher = {Association for Computational Linguistics},
  booktitle = {Proceedings of the 2018 Conference of the North American
               Chapter of the Association for Computational Linguistics:
               Human Language Technologies},
  pages = {(to appear)},
  month = jun,
  year = {2018},
  address = {New Orleans, LA, USA}
}
\end{verbatim}
\twocolumn
\fi

\maketitle

\begin{abstract}
Arguing without committing a fallacy is one of the main requirements of an ideal debate. But even when debating rules are strictly enforced and fallacious arguments punished, arguers often lapse into attacking the opponent by an \emph{ad hominem} argument. As existing research lacks solid empirical investigation of the typology of ad hominem arguments as well as their potential causes, this paper fills this gap by (1) performing several large-scale annotation studies, (2) experimenting with various neural architectures and validating our working hypotheses, such as controversy or reasonableness, and (3) providing linguistic insights into triggers of ad hominem using explainable neural network architectures.
\end{abstract}

\section{Introduction}
\label{sec:introduction}

Human reasoning is lazy and biased but it perfectly serves its purpose in the argumentative context \cite{Mercier.Sperber.2017.book}. When challenged by genuine back-and-forth argumentation, humans do better in both generating and evaluating arguments \cite{Mercier.2011}. The dialogical perspective on argumentation has been reflected in argumentation theory prominently by the pragma-dialectic model of argumentation \cite{vanEemeren.Grootendorst.1992}. Not only sketches this theory an ideal normative model of argumentation but also distinguishes the wrong argumentative moves, \emph{fallacies} \cite{vanEemeren.Grootendorst.1987}. Among the plethora of prototypical fallacies, notwithstanding the controversy of most taxonomies \cite{Boudry.et.al.2015}, \emph{ad hominem} argument is perhaps the most famous one. Arguing \emph{against the person} is considered faulty, yet is prevalent in online and offline discourse.\footnote{According to `Godwin's law' known from the internet pop-culture (\url{https://en.wikipedia.org/wiki/Godwin's_law}), if a discussion goes on long enough, sooner or later someone will compare someone or something to Adolf Hitler.}

Although the ad hominem fallacy has been known since Aristotle, surprisingly there are very few empirical works investigating its properties. While \newcite{Sahlane.2012} analyzed ad hominem and other fallacies in several hundred newspaper editorials, others usually only rely on few examples, as observed by \newcite{DeWijze.2002}. As \newcite{Macagno.2013.ArgJournal} concludes, ad hominem arguments should be considered as multifaceted and complex strategies, involving not a simple argument, but several
combined tactics. However, such research, to the best of our knowledge, does not exist. Very little is known not only about the feasibility of ad hominem theories in practical applications (the NLP perspective) but also about the dynamics and triggers of ad hominem (the theoretical counterpart).

This paper investigates the research gap at three levels of increasing discourse complexity: ad hominem in isolation, direct ad hominem without dialogical exchange, and ad hominem in large inter-personal discourse context. We asked the following research questions. First, what qualitative and quantative properties do ad hominem arguments have in Web debates and how does that reflect the common theoretical view (RQ1)? Second, how much of the debate context do we need for recognizing ad hominem by humans and machine learning systems (RQ2)? And finally, what are the actual triggers of ad hominem arguments and can we predict whether the discussion is going to end up with one (RQ3)?

We tackle these questions by leveraging Web-based argumentation data (\emph{Change my View} on Reddit), performing several large-scale annotation studies, and creating a new dataset. We experiment with various neural architectures and extrapolate the trained models to validate our working hypotheses. Furthermore, we propose a list of potential linguistic and rhetorical triggers of ad hominem based on interpreting parameters of trained neural models.\footnote{An attempt to address the plea for thinking about problems, cognitive science, and the details of human language \cite{Manning.2015.CoLi}.} This article thus presents the first NLP work on multi-faceted ad hominem fallacies in genuine dialogical argumentation. We also release the data and the source code to the research community.\footnote{\url{https://github.com/UKPLab/naacl2018-before-name-calling-habernal-et-al}}

\section{Theoretical background and related work}
\label{sec:related.work}

The prevalent view on argumentation emphasizes its pragmatic goals, such as persuasion and group-based deliberation \cite{vanEemeren.et.al.2014}, although numerous works have dealt with argument as product, that is, treating a single argument and its properties in isolation \cite{Toulmin.1958,Habernal.Gurevych.2017.COLI}. Yet the social role of argumentation and its alleged responsibility for the very skill of human reasoning explained from the evolutionary perspective \cite{Mercier.Sperber.2017.book} provide convincing reasons to treat argumentation as an inherently dialogical tool.

The observation that some arguments are in fact `deceptions in disguise' was made already by Aristotle \cite{Aristotle.1991}, for which the term \emph{fallacy} has been adopted. Leaving the controversial typology of fallacies aside \cite{Hamblin.1970,vanEemeren.Grootendorst.1987,Boudry.et.al.2015}, the \emph{ad hominem} argument is addressed in most theories.
Ad hominem argumentation relies on the strategy of attacking the opponent and some feature of the opponent's character instead of the counter-arguments \cite{Tindale.2007}. With few exceptions, the following five sub-types of ad hominem are prevalent in the literature:
\textbf{abusive ad hominem} (a pure attack on the character of the opponent), 
\textbf{tu quoque ad hominem} (essentially analogous to the ``He did it first'' defense of a three-year-old in a sandbox), \textbf{circumstantial ad hominem} (the ``practice what you preach'' attack and accusation of hypocrisy),
\textbf{bias ad hominem} (the attacked opponent has a hidden agenda), and
\textbf{guilt by association} (associating the opponent with somebody with a low credibility) \cite{Schiappa.Nordin.2013,Macagno.2013.ArgJournal,Walton.2007a,Hansen.2017,Woods.2008}. We omit examples here as these provided in theoretical works or textbooks are usually artificial, as already criticized by \cite{DeWijze.2002} or \cite{Boudry.et.al.2015}.

The topic of fallacies, which might be considered as sub-topic of argumentation quality, has recently been investigated also in the NLP field. Existing works are, however, limited to the monological view \cite{Wachsmuth.et.al.2017.ACL,Habernal.Gurevych.2016.ACL,Habernal.Gurevych.2016.EMNLP,Stab.Gurevych.2017.EACL} or they focus primarily on learning fallacy recognition by humans \cite{Habernal.et.al.2017.EMNLP,Habernal.et.al.2018.LREC}.
Another related NLP sub-field includes abusive language and personal attacks in general.
\newcite{Wulczyn.et.al.2017.WWW} investigated whether or not Wikipedia talk page comments are personal attacks and annotated 38k instances resulting in a highly skewed distribution (only 0.9\% were actual attacks). Regarding the participants' perspective, \newcite{Jain.et.al.2014.LREC} examined principal roles in 80 discussions from the \emph{Wikipedia: Article for Deletion} pages (focusing on stubbornness or ignoredness, among others) and found several typical roles, including `rebels', `voices', or `idiots'. In contrast to our data under investigation (Change My View debates), Wikipedia talk pages do not adhere to strict argumentation rules with manual moderation and have a different pragmatic purpose.

Reddit as a source platform has also been used in other relevant works. \newcite{Saleem.et.al.2016.LREC.WS} detected hateful speech on Reddit by exploiting particular sub-communities to automatically obtain training data.
\newcite{Wang.et.al.2016.NLP.SocSci.WS} experimented with an unsupervised neural model to cluster social roles on sub-reddits dedicated to computer games.
\newcite{Zhang.et.al.2017.ICWSM} proposed a set of nine comment-level dialogue act categories and annotated 9k threads with 100k comments and built a CRF classifier for dialogue act labeling.
Unlike these works which were not related to argumentation, \newcite{Tan.2016} examined persuasion strategies on Change My View using word overlap features.
In contrast to our work, they focused solely on the successful strategies with delta-awarded posts.
Using the same dataset, \newcite{Musi.2017} recently studied concession in argumentation.

\section{Data}
\label{sec:data}

\emph{Change My View} (CMV) is an online `place to post an opinion you accept [...] in an effort to understand other perspectives on the issue', in other words an online platform for `good-faith' argumentation hosted on Reddit.\footnote{\url{https://www.reddit.com/r/changemyview/}} A user posts a \textbf{submission} (also called \textbf{original post(er); OP}) and other participants provide arguments to change the OP's view, forming a typical tree-form Web discussion. A special feature of CMV is that the OP acknowledges convincing arguments by giving a \textbf{delta} point ($\Delta$). Unlike the vast majority of internet discussion forums, CMV enforces obeying strict rules (such as no `low effort' posts, or accusing of being unwilling to change view) whose violation results into deleting the comment by moderators.
These formal requirements of an ideal debate with the notion of violating rules correspond to incorrect moves in critical discussion in the normative pragma-dialectic theory \cite{vanEemeren.Grootendorst.1987}. \emph{Thus, violating the rule of `not being rude or hostile' is equivalent to committing ad hominem fallacy.}
For our experiments, we scraped, in cooperation with Reddit, the complete CMV including the content of the deleted comments so we could fully reconstruct the fallacious discussions, relying on the rule violation labels provided by the moderators. The dataset contains $\approx$ 2M posts in 32k submissions, forming 780k unique threads.

\begin{figure}
\centering
\includegraphics[width=0.8\columnwidth]{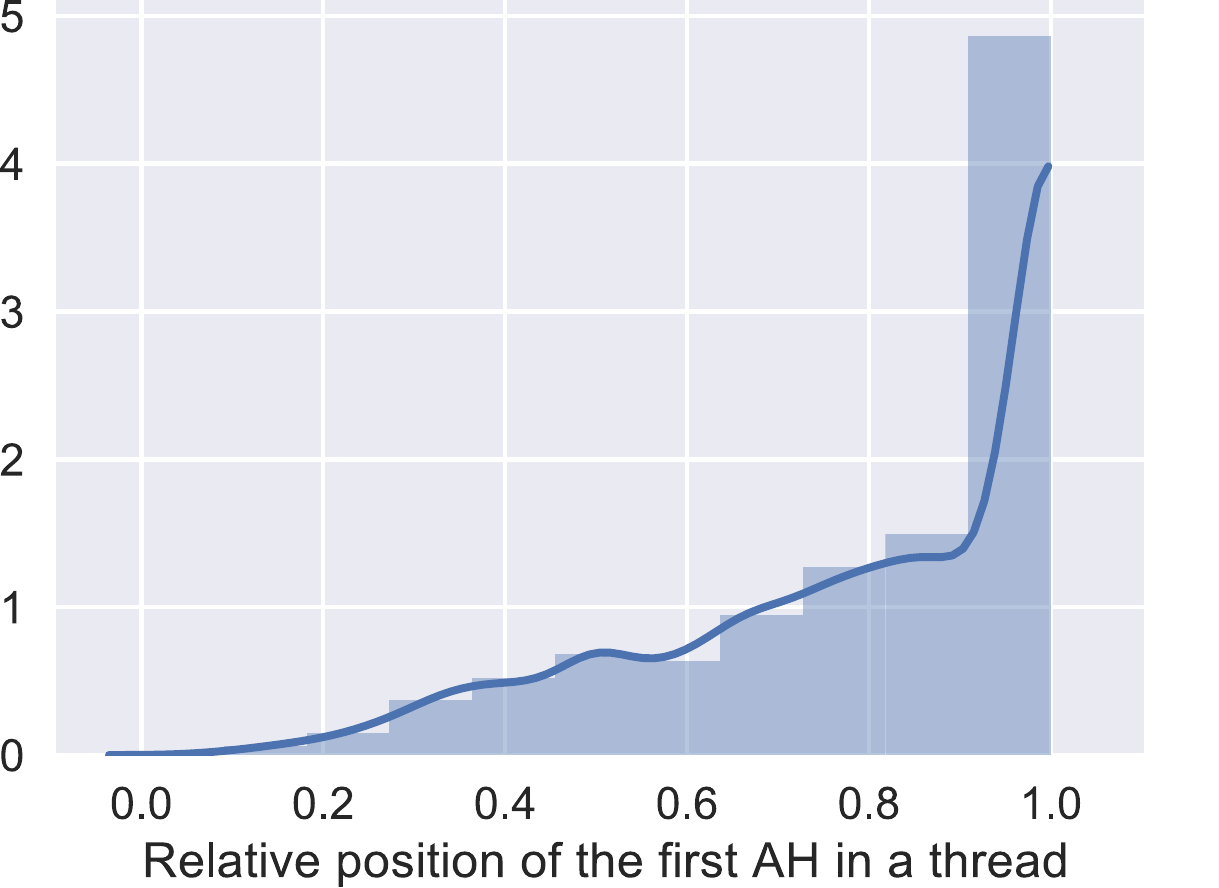}
\caption{\label{fig:before-ah-thread} `No discussion after ad hominem.' Distribution of the number of comments before the first ad hominem is committed proportional to the thread length.}
\end{figure}

We will set up the stage for further experiments by providing several quantitative statistics we performed on the dataset. Only 0.2\% posts in CMV are ad hominem arguments. This contrasts with a typical online discussion: \newcite{Coe.et.al.2014} found 19.5\% of comments under online news articles to be incivil. Most threads contain only a single ad hominem argument (3,396 threads; there are 3,866 ad hominem arguments in total in CMV); only 35 threads contain more than three ad hominem arguments. In 48.6\% of threads containing a single ad hominem, the ad hominem argument is the very last comment. This corresponds to the popular belief that if one is out of arguments, they start attacking and the discussion is over. This trend is also shown in Figure \ref{fig:before-ah-thread} which displays the relative position of the first ad hominem argument in a thread. Replying to ad hominem with another ad hominem happens only in 15\% of the cases; this speaks for the attempts of CMV participants to keep up with the standards of a rather rational discussion.

Regarding ad hominem authors, about 66\% of them start attacking `out of blue', without any previous interaction in the thread. On the other hand, 11\% ad hominem authors write at least one `normal' argument in the thread (we found one outlier who committed ad hominem after writing 57 normal arguments in the thread).
Only in 20\% cases, the ad hominem thread is an interplay between the original poster and another participant. It means that there are usually more people involved in an ad hominem thread. Unfortunately, sometimes the OP herself also commits ad hominem (12\%).

We also investigated the relation between the presence of ad hominem arguments and the submission topic. While most submissions are accompanied by only one or two ad hominem arguments (75\% of submissions), there are also extremes with over 50 ad hominem arguments. Manual analysis revealed that these extremes deal with religion, sexuality/gender, U.S. politics (mostly Trump), racism in the U.S., and veganism. We will elaborate on that later in Section \ref{sec:triggers.first.level}.

\section{Experiments}

The experimental part is divided into three parts according to the increasing level of discourse complexity. We first experiment with ad hominem in isolation in section \ref{sec:ad.hominem.in.cmv}, then with direct ad hominem replies to original posts without dialogical exchange in section \ref{sec:triggers.first.level}, and finally with ad hominem in a larger inter-personal discourse context in section \ref{sec:before.calling.names}.

\subsection{Ad hominem without context in CMV}
\label{sec:ad.hominem.in.cmv}

The first experimental set-up examines ad hominem arguments in \emph{Change my view} regardless of its dialogical context.

\subsubsection{Data verification}
\label{sec:data.verification}

Ad hominem arguments labeled by the CMV moderators come with no warranty. To verify their reliability, we conducted the following annotation studies. First, we needed to estimate parameters of crowdsourcing and its reliability. We sampled 100 random arguments from CMV without context: positive candidates were the reported ad hominem arguments, whereas negative candidates were sampled from comments that either violate other argumentation rules or have a delta label. To ensure the maximal content similarity of these two groups, for each positive instance the semantically closest negative instance was selected.\footnote{Similarity was computed using a cosine similarity of average embedding vectors multiplied by the argument length difference to minimize length-related artifacts. The sample was balanced with roughly 50\% positive and 50\% negative instances.} We then experimented with different numbers of Amazon Mechanical Turk workers and various thresholds of the MACE gold label estimator \cite{Hovy.et.al.2013}; comparing two groups of six workers each and 0.9 threshold yielded almost perfect inter-annotator agreement (0.79 Cohen's $\kappa$). We then used this setting (six workers, 0.9 MACE threshold) to annotate another 452 random arguments sampled in the same way as above.

Crowdsourced `gold' labels were then compared to the original CMV labels (balanced binary task: positive instances (ad hominem) and negative instances) reaching accuracy of 0.878. This means that the ad hominem labels from CMV moderators are quite reliable. Manual error analysis of disagreements revealed 11 missing ad hominem labels. These were not spotted by the moderators but were annotated as such by crowd workers.

\subsubsection{Recognizing ad hominem arguments}
\label{sec:recognizing.ah}

We sampled a larger balanced set of positive instances (ad hominem) and negative instances using the same methodology as in section \ref{sec:data.verification}, resulting in 7,242 instances, and casted the task of recognition of ad hominem arguments as a binary supervised task. We trained two neural classifiers, namely a 2-stacked bi-directional LSTM network \cite{Graves.Schmidthuber.2005}, and a convolutional network \cite{Kim.2014}, and evaluated them using 10-fold cross validation. Throughout the paper we use pre-trained \texttt{word2vec} word embeddings \cite{Mikolov.2013}. Detailed hyperparameters are described in the source codes (link provided in section \ref{sec:introduction}). As results in Table \ref{tab:ah-prediction} show, the task of recognizing ad hominem arguments is feasible and almost achieves the human upper bound performance.

\begin{table}
\centering
\begin{small}
\begin{tabular}{lr}
\textbf{Model} & \textbf{Accuracy} \\ \hline
Human upper bound estimate & 0.878 \\
2 Stacked Bi-LSTM & 0.782 \\
CNN & \textbf{0.810}
\end{tabular}
\end{small}
\caption{\label{tab:ah-prediction} Prediction of ad hominem arguments}
\end{table}

\subsubsection{Typology of ad hominem}
\label{sec:typology.of.ah}

While binary classification of ad hominem as presented above might be sufficient for the purpose of red-flagging arguments, theories provide us with a much finer granularity (recall the typology in section \ref{sec:related.work}). To validate whether this typology is empirically relevant, we executed an annotation experiment to classify ad hominem arguments into the provided five types (plus `other' if none applies). We sampled 200 ad hominem arguments from threads in which interlocution happens only between two persons and which end up with ad hominem. The Mechanical Turk workers were shown this last ad hominem argument as well as the preceding one. Each instance was annotated by 16 workers to achieve a stable distribution of labels as suggested by \newcite{Aroyo.Welty.2015}. While 41\% arguments were categorized as \emph{abusive}, other categories (\emph{tu quoque}, \emph{circumstantial}, and \emph{guilt by association}) were found to be rather ambiguous with very subtle differences. In particular, we observed a very low percentage agreement on these categories and a label distribution spiked around two or more categories. After a manual inspection we concluded that (1) the theoretical typology does not account for longer ad hominem arguments that mix up different attacks and that (2) there are actual phenomena in ad hominem arguments not covered by theoretical categories. These observations reflect those of \newcite[p.~399]{Macagno.2013.ArgJournal} about ad hominem moves as multifaceted strategies.

We thus propose a list of phenomena typical to ad hominem arguments in CMV based on our empirical study. For this purpose, we follow up with another annotation experiment on 400 arguments, with seven workers per instance.\footnote{Here we decided on seven workers per item by relying on other span annotation experiments done in a similar setup \cite{Habernal.et.al.2018.NAACL.Reasoning}.} The goal was to annotate a text span which made the argument an ad hominem; a single argument could contain several spans. We estimated the gold spans using MACE and performed a manual post-analysis by designing a typology of causes of ad hominem together with their frequency of occurrence. The results and examples are summarized in Table \ref{tab:typology-spans}.

\begin{table*}[h!]
\begin{footnotesize}
{\renewcommand{\arraystretch}{1.25}%
\begin{tabular}{p{13em}rp{32em}}
\textbf{Type} &\textbf{ (\%)} & \textbf{Example spans} \\ \hline
Vulgar insult & 31.3 & "Your just an asshole", "you dumb fuck", etc. \\ \hline
Illiteracy insult & 13.0 & "Reading comprehension is your friend", "If you can't grasp the concept, I can't help you" \\ \hline
Condescension & 6.5 & "little buddy", "sir", "boy", "Again, how old are you?" \\ \hline
Ridiculing and sarcasm & 6.5 & "Thank you so much for all your pretentious explanations", "Can you also use Google?" \\ \hline
`Idiot'-insults & 6.5 & "Ever have discussions with narcissistic idiots on the internet? They are so tiring" \\ \hline
Accusation of stupidity & 4.3 & "You have no capability to understand why", "You're obviously just Nobody with enough brains to operate a computer could possibly believe something this stupid" \\ \hline
Lack of argumentation skills & 4.3 & "You're making the claims, it's your job to prove it. Don't you know how debating works?", "You're trash at debating." \\ \hline
Accusation of trolling & 3.9 & "You're just a dishonest troll", "You're using troll tactics" \\ \hline
Accusation of ignorance & 3.5 & "Please dont waste peoples time pretending to know what you're talking about", "Do you even know what you're saying?" \\ \hline
"You didn't read what I wrote" & 3.0 & "Read what I posted before acting like a pompous ass", "Did you even read this?" \\ \hline
"What you say is idiotic" & 2.6 & "To say that people intrinsically understand portion size is idiotic.", "Your second paragraph is fairly idiotic" \\ \hline
Accusation of lying & 2.6 & "Possible lie any harder?", "You are just a liar." \\ \hline
"You don't face the facts and ignore the obvious" & 1.7 & "Willful ignorance is not something I can combat", "How can you explain that? You can't because it will hurt your feelings to face reality" \\ \hline
Accusation of ad hominem or other fallacies & 1.7 & "You started with a fallacy and then deflected.", "You still refuse to acknowledge that you used a strawman argument against me" \\ \hline
Other & 8.3 & "Wow. Someone sounds like a bit of an anti-semite", "You're too dishonest to actually quote the verse because you know it's bullshit"
\end{tabular}
}
\end{footnotesize}
\caption{\label{tab:typology-spans} What makes an argument ad hominem: results of the empirical study of labeling spans in 400 ad hominem arguments.}
\end{table*}

\subsubsection{Results and interpretation}

The data verification annotation study (section \ref{sec:data.verification}) has two direct consequences. First, the high $\kappa$ score (0.79) answers RQ2: for recognizing ad hominem argument, no previous context is necessary. Second, we still found 5\% overlooked ad hominem arguments in CMV thus a moderation-facilitating tool might come handy; this can be served by the well-performing CNN model (0.810 accuracy; section \ref{sec:recognizing.ah}).

The existing theoretical typology of ad hominem arguments, as presented for example in most textbooks, provides only a very simplified view. On the one hand, some of the categories which we found in the empirical labeling study (section \ref{sec:typology.of.ah}) do map to their corresponding counterparts (such as the vulgar insults). On the other hand, some ad hominem insults typical to online argumentation (illiteracy insults, condescension) are not present in studies on ad hominem. Hence, we claim that any potential typology of ad hominem arguments should be multinomial rather than categorical, as we found multiple different spans in a single argument.

\subsection{Triggers of first level ad hominem}
\label{sec:triggers.first.level}

In the following section, we increase the complexity of the studied discourse by taking the original post into account.

\subsubsection{Annotation study}

We already showed that ad hominem arguments are usually preceded by a discussion between the interlocutors. However, 897 submissions (original posts; OPs) have at least one intermediate ad hominem (in other words, the original post is directly attacked). We were thus interested in what triggers these first-level ad hominem arguments. We hypothesize two causes: (1) the \emph{controversy} of the OP, similarly to some related works on news comments \cite{Coe.et.al.2014} and (2) the \emph{reasonableness} of the OP (whether the topic is reasonable to argue about). We model both features on a three-point scale, namely \emph{controversy}: 1 = `not really controversial', 2 = `somehow controversial', 3 = `very controversial' and \emph{reasonableness}: 1 = `quite stupid', 2 = `neutral', 3 = `quite reasonable'.\footnote{Examples of not really controversial: \emph{"I Don't Think Monty Python is Funny"}, very controversial: \emph{"Blacks are generally intellectual inferior to the other major races"}, quite stupid: \emph{"Burritos are better than sandwiches"}, and quite reasonable: \emph{"Nations whose leadership is based upon religion are fundamentally backwards"}.}

We sampled two groups of OPs: those which had some ad hominem arguments in any of its threads but no delta (ad hominem group) and those without ad hominem but some deltas (Delta group). In total, 1,800 balanced instances were annotated by five workers and the resulting value was averaged for each item.\footnote{A pilot crowd sourcing annotation with 5 + 5 workers showed a fair reliability for controversy (Spearman's $\rho$ 0.804) and medium reliability for reasonableness (Spearman's $\rho$ 0.646).}

Statistical analysis of the annotated 1,800 OPs revealed that ad hominem arguments are associated with more controversial OPs (mean controversy 1.23) while delta-awarded arguments with less controversial OPs (mean controversy 1.06; K-S test;\footnote{Kolmogorov-Smirnov (K-S) test is a non-parametric test without any assumptions about the underlying probability distribution.} statistics 0.13, P-value: $7.97\times10^{-7}$).
On the other hand, reasonableness does not seem to play such a role. The difference between ad hominem in reasonable OPs (mean 1.20) and delta in reasonable OPs (mean 1.11) is not that statistically strong; (K-S test statistics: 0.07, P-value: 0.02).

\subsubsection{Regression model for predicting controversy and reasonableness}

We further built a regression model for predicting controversy and reasonableness of the OPs. Along with Bi-LSTM and CNN networks (same models as in \ref{sec:recognizing.ah}) we also developed a neural model that integrates CNN with topic distribution (CNN+LDA). The motivation for a topic-incorporating model was based on our earlier observations presented in section \ref{sec:data}. In particular, we trained an LDA topic model ($k$ = 50) \cite{Blei.et.al.2003} on the heldout OPs and during training/testing, we merged the estimated topic distribution vector with the output layer after convolution and pooling. We performed 10-fold cross validation on the 1,800 annotated OPs and got reasonable performance for controversy prediction ($\rho$ 0.569) and medium performance for reasonableness prediction ($\rho$ 0.385), respectively; both using the CNN+LDA model (see Table \ref{tab:controversy-reasonableness}).

\begin{table}
\centering
\begin{small}
\begin{tabular}{lr}
\multicolumn{2}{c}{\textbf{Controversy (Spearman's $\rho$)}} \\ \hline
Human upper bounds & 0.804 \\
Bi-LSTM & 0.539 \\
CNN	& 0.559 \\
CNN-LDA	& \textbf{0.569} \\
\multicolumn{2}{c}{\textbf{Reasonableness (Spearman's $\rho$)}}  \\ \hline
Human upper bounds & 0.646 \\
Bi-LSTM	& 0.332 \\
CNN	& 0.320 \\
CNN-LDA & \textbf{0.385}
\end{tabular}
\end{small}
\caption{\label{tab:controversy-reasonableness} Results of predicting controversy and reasonableness of the original post.}
\end{table}

We then used the trained model and extrapolated on all held-out OPs (1,267 ad hominem and 10,861 delta OPs, respectively). The analysis again showed that ad hominem arguments tend to be found under more controversial OPs whereas delta arguments in the less controversial ones (K-S test statistics: 0.14, P-value: $1\times10^{-18}$). For reasonableness, the rather low performance of the predictor does not allow us draw any conclusions on the extrapolated data.

\subsubsection{Results and interpretation}

Controversy of the original post is immediately heating up the debate participants and correlates with a higher number of direct ad hominem responses. This corresponds to observations made in comments in newswire where `weightier' topics tended to stir incivility \cite{Coe.et.al.2014}. On the other hand, `stupidity' (or `reasonableness') does not seem to play any significant role. The CNN+LDA model for predicting controversy ($\rho$ 0.569) might come handy for signaling potentially `heated' discussions.

\subsection{Before calling names}
\label{sec:before.calling.names}

In this section, we focus on the dialogical aspect of CMV debates and dynamics of ad hominem fallacies. Although ad hominem arguments appear in many forms (Section \ref{sec:typology.of.ah}), we treat all ad hominem arguments equal in the following experiments.

\subsubsection{Data sampling}
So far we explored what makes an ad hominem argument and whether debated topic influences the number of intermediate attacks. However, possible causes of the argumentative dynamics that ends up with an ad hominem argument remain an open question, which has been addressed in neither argumentation theory nor in cognitive psychology, to the best of our knowledge. We thus cast an explanation of triggers and dynamics of ad hominem discussions as a supervised machine learning problem and draw theoretical insights by a retrospective interpretation of the learned models.

\begin{figure}
\centering
\includegraphics[width=0.99\columnwidth]{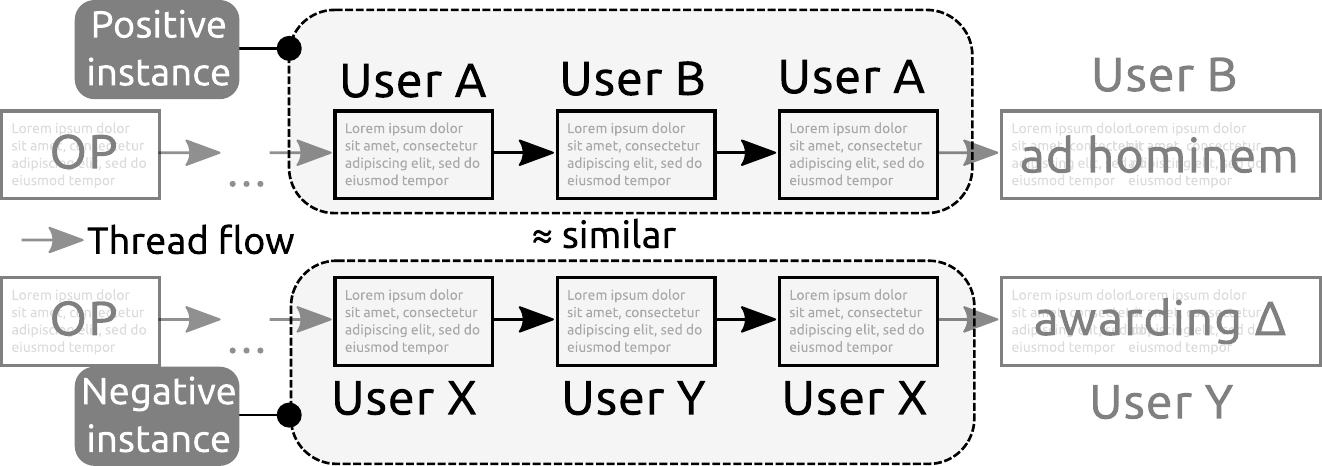}
\caption{\label{fig:thread-sampling} Sampling instances for learning triggers of ad hominem.}
\end{figure}

We sample positive instances by taking three contextual arguments preceding the ad hominem argument from threads which are an interplay between two persons. Negative samples are drawn similarly from threads in which the argument is awarded with $\Delta$ as shown in Figure \ref{fig:thread-sampling}.\footnote{To ensure as much content similarity as possible, we used the same similarity sampling as in section \ref{sec:data.verification}.} Each instance consists of the three concatenated arguments delimited by a special OOV token. This resulted in 2,582 balanced training instances.

\subsubsection{Neural models}

The alleged lack of interpretability of neural networks has motivated several lines of approaches, such as layer-wise relevance propagation \cite{Arras.et.al.2017.WASSA} or representation erasure \cite{Li.et.al.2016.arXiv}, both on sentiment analysis.
As our task at hand deals with multi-party discourse that presumably involves temporal relations important for the learned representation, we opted for a state-of-the-art self-attentive LSTM model. In particular, we re-implemented the Structured Self-Attentive Embedding Neural Network (SSAE-NN) \cite{Lin.et.al.2017.ICLR} which learns an embedding matrix representation of the input using attention weights. To make the attention even more interpretable, we replaced the final non-linear MLP layers with a single linear classifier (softmax). By summing over one dimension of the attention embedding matrix, each word from the input sequence gets associated with a single attention weight that gives us insights into the classifier's `features' (still indirectly, as the true representation is a matrix; see the original paper).\footnote{We also experimented with regularizing the attention matrix as the authors proposed, but it resulted in worse performance.} The learning objective is to recognize whether the thread ends up in an ad hominem argument or a delta point. We trained the model in 10-fold cross-validation and although our goal is not to achieve the best performance but rather to gain insight, we also tested a CNN model (accuracy 0.7095) which performed slightly worse than the SSAE-NN model (accuracy 0.7208).

\subsubsection{Results and interpretation}
\label{sec:before.calling.names:results}

During testing the model, we projected attention weights to the original texts as heat maps and manually analyzed 191 true positives (ad hominem threads recognized correctly), as well as 77 false positives (ad hominem threads misclassified as delta) and 84 false negatives (delta as ad hominem), in total about 120k tokens. The full output is available in the supplementary materials, we use IDs as a reference in the following text.

\begin{figure*}

\begin{small}
\textbf{587\_ah\_t1\_cm7djx3}

\noindent(OOV\_comment\_begin) If only you would n't rely on [ \colorbox[rgb]{0.36,0.42,0.95}{\strut \textcolor{white}{fallacious}} \colorbox[rgb]{0.79,0.80,0.95}{\strut ]} ( http : OOV ) [ \colorbox[rgb]{0.82,0.84,0.95}{\strut arguments} ] ( http : OOV ) to make your point. \colorbox[rgb]{0.81,0.82,0.95}{\strut So} \colorbox[rgb]{0.82,0.83,0.95}{\strut no} \colorbox[rgb]{0.77,0.79,0.95}{\strut ,} \colorbox[rgb]{0.53,0.57,0.95}{\strut \textcolor{white}{I}} \colorbox[rgb]{0.81,0.83,0.95}{\strut do} \colorbox[rgb]{0.77,0.79,0.95}{\strut n't} \colorbox[rgb]{0.74,0.76,0.95}{\strut realize} \colorbox[rgb]{0.84,0.85,0.95}{\strut how} \colorbox[rgb]{0.67,0.69,0.95}{\strut stupid} \colorbox[rgb]{0.84,0.85,0.95}{\strut and} \colorbox[rgb]{0.73,0.75,0.95}{\strut naive} \colorbox[rgb]{0.41,0.47,0.95}{\strut \textcolor{white}{I}} \colorbox[rgb]{0.72,0.75,0.95}{\strut am.} \colorbox[rgb]{0.76,0.78,0.95}{\strut All} \colorbox[rgb]{0.00,0.09,0.95}{\strut \textcolor{white}{I}} \colorbox[rgb]{0.55,0.59,0.95}{\strut \textcolor{white}{'ve}} \colorbox[rgb]{0.59,0.63,0.95}{\strut realized} \colorbox[rgb]{0.79,0.81,0.95}{\strut is} \colorbox[rgb]{0.84,0.85,0.95}{\strut that} \colorbox[rgb]{0.77,0.79,0.95}{\strut you} \colorbox[rgb]{0.84,0.85,0.95}{\strut are} \colorbox[rgb]{0.84,0.85,0.95}{\strut n't} actually \colorbox[rgb]{0.84,0.85,0.95}{\strut prepared} to have an \colorbox[rgb]{0.80,0.81,0.95}{\strut actual} \colorbox[rgb]{0.68,0.71,0.95}{\strut discussion} \colorbox[rgb]{0.78,0.79,0.95}{\strut .} 

\noindent\colorbox[rgb]{0.69,0.71,0.95}{\strut (OOV\_comment\_begin)} \colorbox[rgb]{0.74,0.76,0.95}{\strut What} \colorbox[rgb]{0.52,0.56,0.95}{\strut \textcolor{white}{god}} \colorbox[rgb]{0.76,0.78,0.95}{\strut do} \colorbox[rgb]{0.76,0.78,0.95}{\strut you} \colorbox[rgb]{0.65,0.68,0.95}{\strut believe} \colorbox[rgb]{0.77,0.79,0.95}{\strut in} \colorbox[rgb]{0.65,0.68,0.95}{\strut ?} \colorbox[rgb]{0.65,0.68,0.95}{\strut And} \colorbox[rgb]{0.15,0.23,0.95}{\strut \textcolor{white}{it}} \colorbox[rgb]{0.76,0.78,0.95}{\strut 's} \colorbox[rgb]{0.75,0.77,0.95}{\strut not} \colorbox[rgb]{0.71,0.73,0.95}{\strut a} \colorbox[rgb]{0.56,0.60,0.95}{\strut \textcolor{white}{fallacy}} \colorbox[rgb]{0.71,0.73,0.95}{\strut when} \colorbox[rgb]{0.56,0.60,0.95}{\strut \textcolor{white}{it}} \colorbox[rgb]{0.83,0.84,0.95}{\strut 's} \colorbox[rgb]{0.81,0.82,0.95}{\strut very} comparable to the most popular \colorbox[rgb]{0.83,0.84,0.95}{\strut gods} . 

\noindent\colorbox[rgb]{0.77,0.78,0.95}{\strut (OOV\_comment\_begin)} \colorbox[rgb]{0.67,0.70,0.95}{\strut You} \colorbox[rgb]{0.79,0.80,0.95}{\strut 're} \colorbox[rgb]{0.79,0.80,0.95}{\strut making} \colorbox[rgb]{0.81,0.83,0.95}{\strut an} \colorbox[rgb]{0.71,0.73,0.95}{\strut assumption} \colorbox[rgb]{0.78,0.79,0.95}{\strut on} \colorbox[rgb]{0.56,0.60,0.95}{\strut \textcolor{white}{what}} \colorbox[rgb]{0.61,0.65,0.95}{\strut I} \colorbox[rgb]{0.80,0.81,0.95}{\strut believe} , then \colorbox[rgb]{0.54,0.58,0.95}{\strut \textcolor{white}{attacking}} \colorbox[rgb]{0.68,0.71,0.95}{\strut your} \colorbox[rgb]{0.68,0.71,0.95}{\strut assumption} \colorbox[rgb]{0.80,0.81,0.95}{\strut of} \colorbox[rgb]{0.62,0.65,0.95}{\strut what} \colorbox[rgb]{0.74,0.76,0.95}{\strut my} \colorbox[rgb]{0.72,0.74,0.95}{\strut belief} \colorbox[rgb]{0.80,0.82,0.95}{\strut is} \colorbox[rgb]{0.80,0.82,0.95}{\strut without} \colorbox[rgb]{0.78,0.80,0.95}{\strut me} \colorbox[rgb]{0.81,0.82,0.95}{\strut even} \colorbox[rgb]{0.61,0.64,0.95}{\strut telling} \colorbox[rgb]{0.68,0.71,0.95}{\strut you} \colorbox[rgb]{0.60,0.64,0.95}{\strut anything.} \colorbox[rgb]{0.67,0.70,0.95}{\strut OOV} \colorbox[rgb]{0.51,0.55,0.95}{\strut \textcolor{white}{It}} \colorbox[rgb]{0.70,0.72,0.95}{\strut is} \colorbox[rgb]{0.75,0.77,0.95}{\strut a} \colorbox[rgb]{0.72,0.74,0.95}{\strut OOV} \colorbox[rgb]{0.53,0.57,0.95}{\strut \textcolor{white}{It}} \colorbox[rgb]{0.73,0.75,0.95}{\strut 's} \colorbox[rgb]{0.76,0.78,0.95}{\strut the} \colorbox[rgb]{0.38,0.44,0.95}{\strut \textcolor{white}{comparison}} \colorbox[rgb]{0.59,0.63,0.95}{\strut itself} that is \colorbox[rgb]{0.80,0.82,0.95}{\strut OOV} \colorbox[rgb]{0.83,0.84,0.95}{\strut If} they were n't \colorbox[rgb]{0.84,0.85,0.95}{\strut comparable} at all \colorbox[rgb]{0.85,0.86,0.95}{\strut ,} \colorbox[rgb]{0.82,0.84,0.95}{\strut then} \colorbox[rgb]{0.73,0.76,0.95}{\strut it} 'd be impossible to commit the \colorbox[rgb]{0.85,0.86,0.95}{\strut OOV} \colorbox[rgb]{0.73,0.76,0.95}{\strut You} can compare apples to oranges \colorbox[rgb]{0.80,0.82,0.95}{\strut ,} \colorbox[rgb]{0.51,0.56,0.95}{\strut \textcolor{white}{but}} the \colorbox[rgb]{0.84,0.85,0.95}{\strut moment} you use your fingernails to peel an apple you \colorbox[rgb]{0.69,0.71,0.95}{\strut look} like an \colorbox[rgb]{0.72,0.74,0.95}{\strut idiot} . 
\end{small}
\caption{\label{fig:prediction.heat.map} An example of reconstructed word weight heat map extracted from the attention matrix for a thread which ends up in ad hominem; three previous arguments are shown (see Figure \ref{fig:thread-sampling} for sampling details).}
\end{figure*}

In the following analysis, we solely relied on the weights of words or phrases learned by the attention model, see an example in Figure \ref{fig:prediction.heat.map}. Based on our observations, we summarize several linguistic and argumentative phenomena with examples most likely responsible for ad hominem threads in Table \ref{tab:triggers.of.ah}.

The identified phenomena have few interesting properties in common. First, they all are topic-independent rhetorical devices (except for the loaded keywords at the bottom). Second, many of them deal with meta-level argumentation, i.e., arguing about argumentation (such as missing support or fallacy accusations). Third, most of them do not contain profanity (in contrast to the actual ad hominem arguments of which a third are vulgar insults; cf.\ Table \ref{tab:typology-spans}). And finally, all of them should be easy to avoid.

\begin{table*}
\centering
\begin{small}
\begin{tabular}{p{0.2\textwidth}p{0.78\textwidth}}
\textbf{Phenomena} & \textbf{Examples} \\ \hline
Introducing vulgar intensifiers or interrogatives &
\textbf{388(-1)} \emph{``Where the fuck is your idea to ...''},
\textbf{712(-2)} \emph{``no shortage of fucking gun''},
\textbf{1277(-1)} \emph{``This is fucking CMV''},
\textbf{428(-2)} \emph{``I'm fucking trans!''},
\textbf{2018(-2)} \emph{``an arrogant fuck''},
\textbf{1277(-2)} \emph{``What the fuck are you smoking?''} \\ \hline
Direct imperatives & 
\textbf{1003(-3)} \emph{``You should get more mad about it"}, 
\textbf{857(-2)} \emph{``You need to do a lot better than that."}, 
\textbf{233(-2)} \emph{``So now delete your post"}, 
\textbf{749(-1)} \emph{``google this fact as well"}, 
\textbf{1276(-1)} \emph{``Just look back at the reasons why ..."} \\ \hline
Accusing of believing in or using propaganda &
\textbf{522(-1)} \emph{``It's right wing propaganda?"},
\textbf{1003(-1)} \emph{``If you're not outraged, you're not paying attention to our propaganda that says the opposite of literally thousands of published research articles"} \\ \hline
Accusation of fallacies or bad argumentation practice &
\textbf{238(-3)} \emph{``your snide remarks and poor argumentation skills"},
\textbf{1117(-2)} \emph{``you're circle jerking A vs. B"},
\textbf{263(-3)} \emph{``You're grasping at straws"},
\textbf{78(-3)} \emph{``You sure like changing words and statements to make your argument appear more cogent, don't you?"},
\textbf{210(-1)} \emph{``Your arguments range from ... to ..."},
\textbf{1085(-3)} \emph{``It's only a fallacy"},
\textbf{144(-1)} \emph{``You haven't presented any evidence or argument that disagrees with anything I've said."},
\textbf{587(-3)} \emph{``If only you wouldn't rely on fallacious arguments"} \\ \hline
Reinterpreting opponent's positions &
\textbf{982(-1)} \emph{``The fact that you obviously think ... reveals ..."},
\textbf{982(-2)} \emph{``What makes you think I see myself ... ?"},
\textbf{1060(-3)} \emph{``That kind of thinking is ..."},
\textbf{760(-1)} \emph{``If I'm understanding you correctly"},
\textbf{405(-1)} \emph{``... deluded yourself into believing factually incorrect things"}
\textbf{587(-1)} \emph{``You're making an assumption on what I believe, then attacking your assumption of what my belief is without me even telling you anything."} \\ \hline
Accusation of not reading the other party's arguments &
\textbf{586(-1)} \emph{``... me without even reading my ..."},
\textbf{240(-1)} \emph{``You are just reading it wrong."},
\textbf{310(-1)} \emph{``Oh, you're not actually reading my ..."} \\ \hline
Pointing at missing or unsupported evidence and facts &
\textbf{1238(-2)} \emph{``unsupported bullshit as before"},
\textbf{1121(-3)} \emph{``you can't chose your facts"},
\textbf{931(-1)} \emph{``If that's your only argument ..."},
\textbf{486(-2)} \emph{``unsubstantiated statement"},
\textbf{486(-1)} \emph{``unsupported claims"},
\textbf{71(-2)} \emph{``factually correct"},
\textbf{915(-1)} \emph{``But for the sake of argument, your points are pitifully .."},
\textbf{388(-3)} \emph{``Please provide statistics ... It's silly to debate statistics without actual numbers."} \\ \hline
UPPERCASE &
\textbf{1238(-3)} \emph{``NO ONE CLAIMED THAT ... ARE NOT ... AGAINST ..."} \\ \hline
Sarcasm &
\textbf{78(-2)} \emph{``But I'm sure you know best"},
\textbf{310(-1)} \emph{``Have a nice day."},
\textbf{1276(-1)} \emph{``Good luck with that"} \\ \hline
Mentions of trolling & 
\textbf{701(-2)} \emph{``Then you are giving trolls the victory then?"} \\ \hline
Loaded keywords &
\emph{Nazi, rape, racist} \\
\end{tabular}
\end{small}
\caption{\label{tab:triggers.of.ah} Phenomena resulting into ad hominem learned by the SSAE-NN model. The first number is the instance ID (available in the supplementary material), the minus number in parentheses is the position of the argument before the ad hominem.}
\end{table*}

\paragraph{Misleading `features'}

False positives revealed properties that misled the network to classify delta threads as ad hominem threads.

\begin{itemize}

\item These include \textbf{topic words} (such as \emph{racism, blacks, slave, abortion}) which reflects the implicit bias in the data.
\item Actual interest mixed with indifference in \textbf{sarcasm} is also problematic (\textbf{185(-2)} \emph{``That's a very interesting ..."}).
\item Another problematic phenomena is also \textbf{expressed disagreement} (\textbf{678(-2)} \emph{``overheated rhetoric"}, \textbf{203(-2)} \emph{``But I suppose this argument is ..."}, \textbf{230(-2)} \emph{``But I don't think it's quite ..."}, \textbf{938(-1)} \emph{``I disagree too, however~..."}).
\end{itemize}

False negatives were caused basically by presence of many `informative' \textbf{content words} (\textbf{980} \emph{unemployment, quarterly publication, inflation data}, \textbf{474} \emph{actual publications, this experiment, biological ailments, medical doctorate}, \textbf{1214}
\emph{graduate degree, education, health insurance}) and \textbf{misinterpreted sarcasm} (\textbf{285(-1)}
\emph{``Also this is a cute analogy''}).

\section{Conclusion}

In this article, we investigated ad hominem argumentation on three levels of discourse complexity. We looked into qualitative and quantative properties of ad hominem arguments, crowdsourced labeled data, experimented with models for prediction (0.810 accuracy; \ref{sec:recognizing.ah}), and proposed an updated typology of ad hominem properties (\ref{sec:typology.of.ah}). We then looked into the dynamics of argumentation to examine the relation between the quality of the original post and immediate ad hominem arguments (\ref{sec:triggers.first.level}). Finally, we exploited the learned representation of Self-Attentive Embedding Neural Network to search for features triggering ad hominem in one-to-one discussions. We found several categories of rhetorical devices as well as misleading features (\ref{sec:before.calling.names:results}).

There are several points that deserve further investigation. First, we have ignored meta-information of the debate participants, such as their overall activity (i.e., whether they are spammers or trolls). Second, the proposed typology of ad hominem causes has not yet been post-verified empirically. Third, we expect that personality traits of the participants (BIG5) may also play a significant role in the argumentative exchange. We leave these points for future work.

We believe that our findings will help gain better understanding of, and hopefully keep restraining from, ad hominem fallacies in good-faith discussions.

\section*{Acknowledgments}

This work has been supported by the ArguAna Project GU~798/20-1 (DFG), and by the DFG-funded research training group ``Adaptive Preparation of Information form Heterogeneous Sources'' (AIPHES, GRK 1994/1).

\bibliography{bibliography}

\begin{thebibliography}{}
\expandafter\ifx\csname natexlab\endcsname\relax\def\natexlab#1{#1}\fi

\bibitem[{Aristotle and {Kennedy~(translator)}(1991)}]{Aristotle.1991}
Aristotle and George {Kennedy~(translator)}. 1991.
\newblock {\em On Rhetoric: A Theory of Civil Discourse\/}.
\newblock Oxford University Press, USA.

\bibitem[{Aroyo and Welty(2015)}]{Aroyo.Welty.2015}
Lora Aroyo and Chris Welty. 2015.
\newblock \href{https://doi.org/10.1609/aimag.v36i1.2564}{{Truth is a lie:
  Crowd truth and the seven myths of human annotation}}.
\newblock {\em AI Magazine\/} 36(1):15--24.
\newblock
  \href{https://doi.org/10.1609/aimag.v36i1.2564}{https://doi.org/10.1609/aimag.v36i1.2564}.

\bibitem[{Arras et~al.(2017)Arras, Montavon, M{\"{u}}ller, and
  Samek}]{Arras.et.al.2017.WASSA}
Leila Arras, Gr{\'{e}}goire Montavon, Klaus-Robert M{\"{u}}ller, and Wojciech
  Samek. 2017.
\newblock \href{http://www.aclweb.org/anthology/W17-5221}{{Explaining Recurrent
  Neural Network Predictions in Sentiment Analysis}}.
\newblock In {\em Proceedings of the 8th Workshop on Computational Approaches
  to Subjectivity, Sentiment and Social Media Analysis\/}. Association for
  Computational Linguistics, Copenhagen, Denmark, pages 159--168.
\newblock
  \href{http://www.aclweb.org/anthology/W17-5221}{http://www.aclweb.org/anthology/W17-5221}.

\bibitem[{Blei et~al.(2003)Blei, Ng, and Jordan}]{Blei.et.al.2003}
David~M. Blei, Andrew~Y. Ng, and Michael~I. Jordan. 2003.
\newblock \href{http://dl.acm.org/citation.cfm?id=944919.944937}{Latent
  dirichlet allocation}.
\newblock {\em Journal of Machine Learning Research\/} 3:993--1022.
\newblock
  \href{http://dl.acm.org/citation.cfm?id=944919.944937}{http://dl.acm.org/citation.cfm?id=944919.944937}.

\bibitem[{Boudry et~al.(2015)Boudry, Paglieri, and
  Pigliucci}]{Boudry.et.al.2015}
Maarten Boudry, Fabio Paglieri, and Massimo Pigliucci. 2015.
\newblock \href{https://doi.org/10.1007/s10503-015-9359-1}{{The Fake, the
  Flimsy, and the Fallacious: Demarcating Arguments in Real Life}}.
\newblock {\em Argumentation\/} 29(4):431--456.
\newblock
  \href{https://doi.org/10.1007/s10503-015-9359-1}{https://doi.org/10.1007/s10503-015-9359-1}.

\bibitem[{Coe et~al.(2014)Coe, Kenski, and Rains}]{Coe.et.al.2014}
Kevin Coe, Kate Kenski, and Stephen~A. Rains. 2014.
\newblock \href{https://doi.org/10.1111/jcom.12104}{{Online and Uncivil?
  Patterns and Determinants of Incivility in Newspaper Website Comments}}.
\newblock {\em Journal of Communication\/} 64(4):658--679.
\newblock
  \href{https://doi.org/10.1111/jcom.12104}{https://doi.org/10.1111/jcom.12104}.

\bibitem[{de~Wijze(2002)}]{DeWijze.2002}
Stephen de~Wijze. 2002.
\newblock \href{https://doi.org/10.1111/1469-5812.00004}{{Complexity, Relevance
  and Character: Problems with Teaching the "Ad Hominem" Fallacy.}}
\newblock {\em Educational Philosophy and Theory\/} 35(1):31--56.
\newblock
  \href{https://doi.org/10.1111/1469-5812.00004}{https://doi.org/10.1111/1469-5812.00004}.

\bibitem[{Graves and Schmidhuber(2005)}]{Graves.Schmidthuber.2005}
Alex Graves and J\"{u}rgen Schmidhuber. 2005.
\newblock \href{https://doi.org/10.1016/j.neunet.2005.06.042}{{Framewise
  phoneme classification with bidirectional LSTM and other neural network
  architectures}}.
\newblock {\em Neural Networks\/} 18(5):602--610.
\newblock
  \href{https://doi.org/10.1016/j.neunet.2005.06.042}{https://doi.org/10.1016/j.neunet.2005.06.042}.

\bibitem[{Habernal and
  Gurevych(2016{\natexlab{a}})}]{Habernal.Gurevych.2016.EMNLP}
Ivan Habernal and Iryna Gurevych. 2016{\natexlab{a}}.
\newblock \href{http://aclweb.org/anthology/D16-1129}{{What makes a convincing
  argument? Empirical analysis and detecting attributes of convincingness in
  Web argumentation}}.
\newblock In {\em Proceedings of the 2016 Conference on Empirical Methods in
  Natural Language Processing\/}. Association for Computational Linguistics,
  Austin, Texas, pages 1214--1223.
\newblock
  \href{http://aclweb.org/anthology/D16-1129}{http://aclweb.org/anthology/D16-1129}.

\bibitem[{Habernal and
  Gurevych(2016{\natexlab{b}})}]{Habernal.Gurevych.2016.ACL}
Ivan Habernal and Iryna Gurevych. 2016{\natexlab{b}}.
\newblock \href{http://www.aclweb.org/anthology/P16-1150}{{Which argument is
  more convincing? Analyzing and predicting convincingness of Web arguments
  using bidirectional {LS}TM}}.
\newblock In {\em Proceedings of the 54th Annual Meeting of the Association for
  Computational Linguistics (Volume 1: Long Papers)\/}. Association for
  Computational Linguistics, Berlin, Germany, pages 1589--1599.
\newblock
  \href{http://www.aclweb.org/anthology/P16-1150}{http://www.aclweb.org/anthology/P16-1150}.

\bibitem[{Habernal and Gurevych(2017)}]{Habernal.Gurevych.2017.COLI}
Ivan Habernal and Iryna Gurevych. 2017.
\newblock \href{https://doi.org/10.1162/COLI\_a\_00276}{{Argumentation Mining
  in User-Generated Web Discourse}}.
\newblock {\em Computational Linguistics\/} 43(1):125--179.
\newblock
  \href{https://doi.org/10.1162/COLI\_a\_00276}{https://doi.org/10.1162/COLI\_a\_00276}.

\bibitem[{Habernal et~al.(2017)Habernal, Hannemann, Pollak, Klamm, Pauli, and
  Gurevych}]{Habernal.et.al.2017.EMNLP}
Ivan Habernal, Raffael Hannemann, Christian Pollak, Christopher Klamm, Patrick
  Pauli, and Iryna Gurevych. 2017.
\newblock \href{http://www.aclweb.org/anthology/D17-2002}{{Argotario:
  Computational Argumentation Meets Serious Games}}.
\newblock In {\em Proceedings of the 2017 Conference on Empirical Methods in
  Natural Language Processing: System Demonstrations\/}. Association for
  Computational Linguistics, Copenhagen, Denmark, pages 7--12.
\newblock
  \href{http://www.aclweb.org/anthology/D17-2002}{http://www.aclweb.org/anthology/D17-2002}.

\bibitem[{Habernal et~al.(2018{\natexlab{a}})Habernal, Pauli, and
  Gurevych}]{Habernal.et.al.2018.LREC}
Ivan Habernal, Patrick Pauli, and Iryna Gurevych. 2018{\natexlab{a}}.
\newblock {Adapting Serious Game for Fallacious Argumentation to German:
  Pitfalls, Insights, and Best Practices}.
\newblock In {\em Proceedings of the Eleventh International Conference on
  Language Resources and Evaluation (LREC 2018)\/}. European Language Resources
  Association (ELRA), Miyazaki, Japan, page (to appear).

\bibitem[{Habernal et~al.(2018{\natexlab{b}})Habernal, Wachsmuth, Gurevych, and
  Stein}]{Habernal.et.al.2018.NAACL.Reasoning}
Ivan Habernal, Henning Wachsmuth, Iryna Gurevych, and Benno Stein.
  2018{\natexlab{b}}.
\newblock \href{https://arxiv.org/abs/1708.01425}{{The Argument Reasoning
  Comprehension Task: Identification and Reconstruction of Implicit Warrants}}.
\newblock In {\em Proceedings of the 2018 Conference of the North American
  Chapter of the Association for Computational Linguistics: Human Language
  Technologies\/}. Association for Computational Linguistics, New Orleans, LA,
  page (to appear).
\newblock
  \href{https://arxiv.org/abs/1708.01425}{https://arxiv.org/abs/1708.01425}.

\bibitem[{Hamblin(1970)}]{Hamblin.1970}
Charles~L. Hamblin. 1970.
\newblock {\em Fallacies\/}.
\newblock Methuen, London, UK.

\bibitem[{Hansen(2017)}]{Hansen.2017}
Hans Hansen. 2017.
\newblock \href{http://plato.stanford.edu/entries/fallacies/}{Fallacies}.
\newblock In Edward~N. Zalta, editor, {\em The Stanford Encyclopedia of
  Philosophy\/}, Metaphysics Research Lab, Stanford University.
\newblock
  \href{http://plato.stanford.edu/entries/fallacies/}{http://plato.stanford.edu/entries/fallacies/}.

\bibitem[{Hovy et~al.(2013)Hovy, Berg-Kirkpatrick, Vaswani, and
  Hovy}]{Hovy.et.al.2013}
Dirk Hovy, Taylor Berg-Kirkpatrick, Ashish Vaswani, and Eduard Hovy. 2013.
\newblock \href{http://www.aclweb.org/anthology/N13-1132}{Learning whom to
  trust with {MACE}}.
\newblock In {\em Proceedings of NAACL-HLT 2013\/}. Association for
  Computational Linguistics, Atlanta, Georgia, pages 1120--1130.
\newblock
  \href{http://www.aclweb.org/anthology/N13-1132}{http://www.aclweb.org/anthology/N13-1132}.

\bibitem[{Jain et~al.(2014)Jain, Bhatia, Rein, and Hovy}]{Jain.et.al.2014.LREC}
Siddharth Jain, Archna Bhatia, Angelique Rein, and Eduard Hovy. 2014.
\newblock {A Corpus of Participant Roles in Contentious Discussions}.
\newblock In {\em Proceedings of the Ninth International Conference on Language
  Resources and Evaluation (LREC'14)\/}. European Language Resources
  Association (ELRA), Reykjavik, Iceland, pages 1751--1756.

\bibitem[{Kim(2014)}]{Kim.2014}
Yoon Kim. 2014.
\newblock \href{http://www.aclweb.org/anthology/D14-1181}{Convolutional neural
  networks for sentence classification}.
\newblock In {\em Proceedings of the 2014 Conference on Empirical Methods in
  Natural Language Processing (EMNLP)\/}. Association for Computational
  Linguistics, Doha, Qatar, pages 1746--1751.
\newblock
  \href{http://www.aclweb.org/anthology/D14-1181}{http://www.aclweb.org/anthology/D14-1181}.

\bibitem[{Li et~al.(2016)Li, Monroe, and Jurafsky}]{Li.et.al.2016.arXiv}
Jiwei Li, Will Monroe, and Dan Jurafsky. 2016.
\newblock \href{http://arxiv.org/abs/1612.08220}{{Understanding Neural Networks
  through Representation Erasure}}.
\newblock In {\em arXiv preprint\/}.
\newblock
  \href{http://arxiv.org/abs/1612.08220}{http://arxiv.org/abs/1612.08220}.

\bibitem[{Lin et~al.(2017)Lin, Feng, {Nogueira dos Santos}, Yu, Xiang, Zhou,
  and Bengio}]{Lin.et.al.2017.ICLR}
Zhouhan Lin, Minwei Feng, Cicero {Nogueira dos Santos}, Mo~Yu, Bing Xiang,
  Bowen Zhou, and Yoshua Bengio. 2017.
\newblock \href{http://arxiv.org/abs/1703.03130}{{A Structured Self-attentive
  Sentence Embedding}}.
\newblock In {\em Proceedings of the 5th International Conference on Learning
  Representations (ICLR)\/}. Toulon, France, pages 1--15.
\newblock
  \href{http://arxiv.org/abs/1703.03130}{http://arxiv.org/abs/1703.03130}.

\bibitem[{Macagno(2013)}]{Macagno.2013.ArgJournal}
Fabrizio Macagno. 2013.
\newblock \href{https://doi.org/10.1007/s10503-013-9291-1}{Strategies of
  character attack}.
\newblock {\em Argumentation\/} 27(4):369--401.
\newblock
  \href{https://doi.org/10.1007/s10503-013-9291-1}{https://doi.org/10.1007/s10503-013-9291-1}.

\bibitem[{Manning(2015)}]{Manning.2015.CoLi}
Christopher~D. Manning. 2015.
\newblock \href{https://doi.org/10.1162/COLI\_a\_00239}{{Computational
  Linguistics and Deep Learning}}.
\newblock {\em Computational Linguistics\/} 41(4):701--707.
\newblock
  \href{https://doi.org/10.1162/COLI\_a\_00239}{https://doi.org/10.1162/COLI\_a\_00239}.

\bibitem[{Mercier and Sperber(2011)}]{Mercier.2011}
Hugo Mercier and Dan Sperber. 2011.
\newblock \href{https://doi.org/10.1017/S0140525X10000968}{{Why do humans
  reason? Arguments for an argumentative theory}}.
\newblock {\em The Behavioral and Brain Sciences\/} 34(2):57--74; discussion
  74--111.
\newblock
  \href{https://doi.org/10.1017/S0140525X10000968}{https://doi.org/10.1017/S0140525X10000968}.

\bibitem[{Mercier and Sperber(2017)}]{Mercier.Sperber.2017.book}
Hugo Mercier and Dan Sperber. 2017.
\newblock {\em {The Enigma of Reason}\/}.
\newblock Harvard University Press, Cambridge, MA, USA.

\bibitem[{Mikolov et~al.(2013)Mikolov, Sutskever, Chen, Corrado, and
  Dean}]{Mikolov.2013}
Tomas Mikolov, Ilya Sutskever, Kai Chen, Greg~S Corrado, and Jeff Dean. 2013.
\newblock Distributed representations of words and phrases and their
  compositionality.
\newblock In C.~J.~C. Burges, L.~Bottou, M.~Welling, Z.~Ghahramani, and K.~Q.
  Weinberger, editors, {\em Advances in Neural Information Processing Systems
  26\/}, Curran Associates, Inc., pages 3111--3119.

\bibitem[{Musi(2017)}]{Musi.2017}
Elena Musi. 2017.
\newblock \href{https://doi.org/10.1177/1461445617734955}{{How did you change
  my view? A corpus-based study of concessions' argumentative role}}.
\newblock {\em Discourse Studies\/} page (in press).
\newblock
  \href{https://doi.org/10.1177/1461445617734955}{https://doi.org/10.1177/1461445617734955}.

\bibitem[{Sahlane(2012)}]{Sahlane.2012}
Ahmed Sahlane. 2012.
\newblock \href{https://doi.org/10.1007/s10503-012-9265-8}{Argumentation and
  fallacy in the justification of the 2003~{W}ar on {Iraq}}.
\newblock {\em Argumentation\/} 26(4):459--488.
\newblock
  \href{https://doi.org/10.1007/s10503-012-9265-8}{https://doi.org/10.1007/s10503-012-9265-8}.

\bibitem[{Saleem et~al.(2016)Saleem, Dillon, Benesch, and
  Ruths}]{Saleem.et.al.2016.LREC.WS}
Haji~Mohammad Saleem, Kelly~P Dillon, Susan Benesch, and Derek Ruths. 2016.
\newblock {A Web of Hate: Tackling Hateful Speech in Online Social Spaces}.
\newblock In Guy {De Pauw}, Ben Verhoeven, Bart Desmet, and Els Lefever,
  editors, {\em Proceedings of the First Workshop on Text Analytics for
  Cybersecurity and Online Safety\/}. European Language Resources Association
  (ELRA), Portoro{\v{z}}, Slovenia, pages 1--9.

\bibitem[{Schiappa and Nordin(2013)}]{Schiappa.Nordin.2013}
Edward Schiappa and John~P. Nordin. 2013.
\newblock {\em Argumentation: Keeping Faith with Reason\/}.
\newblock Pearson UK, 1st edition.

\bibitem[{Stab and Gurevych(2017)}]{Stab.Gurevych.2017.EACL}
Christian Stab and Iryna Gurevych. 2017.
\newblock \href{http://www.aclweb.org/anthology/E17-1092}{{Recognizing
  Insufficiently Supported Arguments in Argumentative Essays}}.
\newblock In {\em Proceedings of the 15th Conference of the European Chapter of
  the Association for Computational Linguistics (EACL 2017)\/}. Association for
  Computational Linguistics, pages 980--990.
\newblock
  \href{http://www.aclweb.org/anthology/E17-1092}{http://www.aclweb.org/anthology/E17-1092}.

\bibitem[{Tan et~al.(2016)Tan, Niculae, Danescu-Niculescu-Mizil, and
  Lee}]{Tan.2016}
Chenhao Tan, Vlad Niculae, Cristian Danescu-Niculescu-Mizil, and Lillian Lee.
  2016.
\newblock \href{http://arxiv.org/abs/1602.01103}{{Winning Arguments:
  Interaction Dynamics and Persuasion Strategies in Good-faith Online
  Discussions}}.
\newblock In {\em Proceedings of the 25th International Conference on World
  Wide Web\/}. ACM, Montreal, CA, page (to appear).
\newblock
  \href{http://arxiv.org/abs/1602.01103}{http://arxiv.org/abs/1602.01103}.

\bibitem[{Tindale(2007)}]{Tindale.2007}
Christopher~W. Tindale. 2007.
\newblock {\em Fallacies and Argument Appraisal\/}.
\newblock Cambridge University Press, New York, NY, USA, critical reasoning and
  argumentation edition.

\bibitem[{Toulmin(1958)}]{Toulmin.1958}
Stephen~E. Toulmin. 1958.
\newblock {\em The Uses of Argument\/}.
\newblock Cambridge University Press.

\bibitem[{van Eemeren et~al.(2014)van Eemeren, Garssen, Krabbe,
  Snoeck~Henkemans, Verheij, and Wagemans}]{vanEemeren.et.al.2014}
Frans~H. van Eemeren, Bart Garssen, Erik C.~W. Krabbe, A.~Francisca
  Snoeck~Henkemans, Bart Verheij, and Jean H.~M. Wagemans. 2014.
\newblock {\em Handbook of Argumentation Theory\/}.
\newblock Springer, Berlin/Heidelberg.

\bibitem[{{van Eemeren} and Grootendorst(1987)}]{vanEemeren.Grootendorst.1987}
Frans~H. {van Eemeren} and Rob Grootendorst. 1987.
\newblock \href{https://doi.org/10.1007/BF00136779}{{Fallacies in
  pragma-dialectical perspective}}.
\newblock {\em Argumentation\/} 1(3):283--301.
\newblock
  \href{https://doi.org/10.1007/BF00136779}{https://doi.org/10.1007/BF00136779}.

\bibitem[{van Eemeren and Grootendorst(1992)}]{vanEemeren.Grootendorst.1992}
Frans~H. van Eemeren and Rob Grootendorst. 1992.
\newblock {\em {Argumentation, communication, and fallacies: a
  pragma-dialectical perspective}\/}.
\newblock Lawrence Erlbaum Associates, Inc.

\bibitem[{Wachsmuth et~al.(2017)Wachsmuth, Naderi, Habernal, Hou, Hirst,
  Gurevych, and Stein}]{Wachsmuth.et.al.2017.ACL}
Henning Wachsmuth, Nona Naderi, Ivan Habernal, Yufang Hou, Graeme Hirst, Iryna
  Gurevych, and Benno Stein. 2017.
\newblock \href{http://aclweb.org/anthology/P17-2039}{{Argumentation Quality
  Assessment: Theory vs. Practice}}.
\newblock In {\em Proceedings of the 55th Annual Meeting of the Association for
  Computational Linguistics (Volume 2: Short Papers)\/}. Association for
  Computational Linguistics, Vancouver, Canada, pages 250--255.
\newblock
  \href{http://aclweb.org/anthology/P17-2039}{http://aclweb.org/anthology/P17-2039}.

\bibitem[{Walton(2007)}]{Walton.2007a}
Douglas Walton. 2007.
\newblock {\em Media Argumentation: Dialect, Persuasion and Rhetoric\/}.
\newblock Cambridge University Press.

\bibitem[{Wang et~al.(2016)Wang, Hamilton, and
  Leskovec}]{Wang.et.al.2016.NLP.SocSci.WS}
Alex Wang, William~L. Hamilton, and Jure Leskovec. 2016.
\newblock \href{http://aclweb.org/anthology/W16-5610}{{Learning Linguistic
  Descriptors of User Roles in Online Communities}}.
\newblock In {\em Proceedings of 2016 EMNLP Workshop on Natural Language
  Processing and Computational Social Science\/}. Association for Computational
  Linguistics, Austin, Texas, pages 76--85.
\newblock
  \href{http://aclweb.org/anthology/W16-5610}{http://aclweb.org/anthology/W16-5610}.

\bibitem[{Woods(2008)}]{Woods.2008}
John Woods. 2008.
\newblock \href{https://doi.org/10.22329/il.v27i1.467}{{Lightening up on the Ad
  Hominem}}.
\newblock {\em Informal Logic\/} 27(1):109.
\newblock
  \href{https://doi.org/10.22329/il.v27i1.467}{https://doi.org/10.22329/il.v27i1.467}.

\bibitem[{Wulczyn et~al.(2017)Wulczyn, Thain, and
  Dixon}]{Wulczyn.et.al.2017.WWW}
Ellery Wulczyn, Nithum Thain, and Lucas Dixon. 2017.
\newblock \href{https://doi.org/10.1145/3038912.3052591}{{Ex Machina: Personal
  Attacks Seen at Scale}}.
\newblock In {\em Proceedings of the 26th International Conference on World
  Wide Web\/}. International World Wide Web Conferences Steering Committee,
  Perth, Australia, pages 1391--1399.
\newblock
  \href{https://doi.org/10.1145/3038912.3052591}{https://doi.org/10.1145/3038912.3052591}.

\bibitem[{Zhang et~al.(2017)Zhang, Culbertson, and
  Paritosh}]{Zhang.et.al.2017.ICWSM}
Amy Zhang, Bryan Culbertson, and Praveen Paritosh. 2017.
\newblock {Characterizing Online Discussion Using Coarse Discourse Sequences}.
\newblock In {\em Proceedings of the Eleventh International AAAI Conference on
  Web and Social Media (ICWSM 2017)\/}. AAAI Press, Montreal, Canada, pages
  357--366.

\end{thebibliography}
\bibliographystyle{aclnatbib}

\end{document}